\documentclass[10pt,journal,onecolumn]{IEEEtran}

\usepackage[utf8]{inputenc} 
\usepackage[T1]{fontenc}    
\usepackage{hyperref}       
\usepackage{url}            
\usepackage{booktabs}       
\usepackage{amsfonts}       
\usepackage{nicefrac}       
\usepackage{microtype}      
\usepackage{amsfonts,amsmath,amsthm,amssymb}       
\usepackage[usenames,dvipsnames]{xcolor}
\usepackage{graphicx}
\usepackage{booktabs}
\usepackage{floatrow}

\newfloatcommand{capbtabbox}{table}[][\FBwidth]

\usepackage{graphicx}
 \usepackage{subcaption}
\usepackage{amsfonts,amssymb,amsmath,amsthm}
\usepackage{color}
\usepackage{url}
\usepackage{tablefootnote}
\usepackage{mathrsfs}
\usepackage{enumitem}
\usepackage{array}
\usepackage{algorithm,algpseudocode}
\usepackage{wrapfig}
\usepackage{xcolor}
\usepackage{mathtools}

\usepackage{nicefrac}
\usepackage{eufrak}
\usepackage{yfonts}

\newlength\myindent
\newcommand{\argmax}{\operatornamewithlimits{argmax}}
\newcommand{\argmin}{\operatornamewithlimits{argmin}}

\newcommand{\ex}[2]{\underset{#1}{\Ebb}\left(#2\right)}

\def\x{\boldsymbol{x}}

\def\z{\boldsymbol{z}}

\def\r{\boldsymbol{r}}

\def\v{\boldsymbol{v}}
\def\p{\boldsymbol{p}}

\def\u{\boldsymbol{u}}
\def\S{\mathcal{S}}

\def\Ebb{\mathbb{E}}

\makeatletter
\newtheorem*{rep@theorem}{\rep@title}
\newcommand{\newreptheorem}[2]{%
\newenvironment{rep#1}[1]{%
 \def\rep@title{#2 \ref{##1}}%
 \begin{rep@theorem}}%
 {\end{rep@theorem}}}
\makeatother

\newreptheorem{theorem}{Theorem}
\newreptheorem{lemma}{Lemma}
\newreptheorem{corollary}{Corollary}

\DeclarePairedDelimiterX{\infdivx}[2]{}{}{%
  #1\;\delimsize\|\;#2%
}

\title{Classification regions of deep neural networks}

\author{
\IEEEauthorblockN{Alhussein Fawzi\IEEEauthorrefmark{1}\IEEEauthorrefmark{3}\thanks{\IEEEauthorrefmark{1}The first two authors contributed equally to this work.}\;\,\thanks{\IEEEauthorrefmark{3}UCLA Vision Lab, University of California, Los Angeles, CA 90095}}
  \IEEEauthorblockA{\texttt{fawzi@cs.ucla.edu}}
  \\
\IEEEauthorblockN{Seyed-Mohsen Moosavi-Dezfooli\IEEEauthorrefmark{1}\IEEEauthorrefmark{2}\;\,\thanks{\IEEEauthorrefmark{2}LTS4, \'Ecole Polytechnique F\'ed\'erale de Lausanne, Switzerland}}
\IEEEauthorblockA{\texttt{seyed.moosavi@epfl.ch}}
\\
  \IEEEauthorblockN{Pascal Frossard\IEEEauthorrefmark{2}}
  \IEEEauthorblockA{\texttt{pascal.frossard@epfl.ch}}
  \\
  \IEEEauthorblockN{Stefano Soatto\IEEEauthorrefmark{3}}
  \IEEEauthorblockA{\texttt{soatto@cs.ucla.edu}}
}

\begin{document}

\maketitle
\begin{abstract}
The goal of this paper is to analyze the geometric properties of deep neural network classifiers in the input space. We specifically study the topology of classification regions created by deep networks, as well as their associated decision boundary.
Through a systematic empirical investigation, we show that state-of-the-art deep nets learn connected classification regions, and that the decision boundary in the vicinity of datapoints is flat along most directions. We further draw an essential connection between two seemingly unrelated properties of deep networks: their sensitivity to additive perturbations in the inputs, and the curvature of their decision boundary. The directions where the decision boundary is curved in fact characterize the directions to which the classifier is the most vulnerable.
We finally leverage a fundamental \textit{asymmetry} in the curvature of the decision boundary of deep nets, and propose a method to discriminate between original images, and images perturbed with small adversarial examples. We show the effectiveness of this purely geometric approach for detecting small adversarial perturbations in images, and for recovering the labels of perturbed images.
\end{abstract}

\section{Introduction}

While the geometry of classification regions and decision functions induced by traditional classifiers (such as linear and kernel SVM) is fairly well understood, these fundamental geometric properties are to a large extent unknown for state-of-the-art deep neural networks. Yet, to understand the recent success of deep neural networks and potentially address their weaknesses (such as their instability to perturbations \cite{szegedy2013intriguing}), an understanding of these geometric properties remains primordial. While many fundamental properties of deep networks have recently been studied, such as their \textit{optimization landscape} in \cite{choromanska2014loss, dauphin2014identifying}, their \textit{generalization} in \cite{zhang2016understanding, hardt2015train}, and their \textit{expressivity} in \cite{delalleau2011shallow, cohen2016convolutional}, the geometric properties of the decision boundary and classification regions of deep networks has comparatively received little attention. The goal of this paper is to analyze these properties, and leverage them to improve the robustness of such classifiers to perturbations. 

In this paper, we specifically view classification regions as topological spaces, and decision boundaries as hypersurfaces and examine their geometric properties. We first study the classification regions induced by state-of-the-art deep networks, and provide empirical evidence suggesting that these classification regions are \textit{connected}; that is, there exists a continuous path that remains in the region between any two points of the same label. Up to our knowledge, this represents the first instance where the connectivity of classification regions is empirically shown.
Then, to study the complexity of the functions learned by the deep network, we analyze the curvature of their decision boundary. We empirically show that
\begin{itemize}
\item The decision boundary in the vicinity of natural images is flat in most directions, with only a very few directions that are significantly curved.
\item We reveal the existence of a fundamental asymmetry in the decision boundary of deep networks, whereby the decision boundary (near natural images) is biased towards negative curvatures.
\item Directions with curved decision boundaries are shared between different datapoints.
\item We demonstrate the existence of a relation between the sensitivity of a classifier to perturbations of the inputs, and these shared directions: a deep net is vulnerable to perturbations along these directions, and is insensitive to perturbations along the remaining directions.
\end{itemize}
We finally leverage the fundamental asymmetry of deep networks revealed in our analysis, and propose an algorithm to detect natural images from imperceptibly similar images with very small adversarial perturbations \cite{szegedy2013intriguing}, as well as estimate the correct label of these  perturbed samples. We show that our purely geometric characterization of (small) adversarial examples is very effective to recognize perturbed samples. 

\textbf{Related works.} In \cite{poole2016exponential}, the authors employ tools from Riemannian geometry to study the expressivity of random deep neural networks. In particular, the largest principal curvatures are shown to increase exponentially with the depth; the decision boundaries hence become more complex with depth. We provide in this paper a complementary and more global analysis of the decision boundary, where the curvature of the decision boundary along \textit{all} directions are analyzed. The authors of  \cite{montufar2014number} show that the number of linear regions (in the input space) of deep networks grow exponentially with the number of layers. 
Note also that unlike \cite{choromanska2014loss, dauphin2014identifying, chaudhari2016entropy, dinh2017sharp} that study the geometry of the optimization function in the \textit{weight space}, we focus here on geometric properties in the input space.
Finally, we note that graph-based techniques have been proposed in \cite{melnik2000using, aupetit2003high} to analyze the classification regions of shallow neural networks; we focus here on the new generation of deep neural networks, which have shown remarkable performance.
\section{Definitions and notations}

Let $f: \mathbb{R}^d \rightarrow \mathbb{R}^L$ denote a $L$ class classifier. Given a datapoint $\x_0 \in \mathbb{R}^d$, the estimated label is obtained by $\hat{k}(\x_0) = \argmax_{k} f_k(\x_0)$, where $f_{k} (\x)$ is the $k^{\text{th}}$ component of $f(\x)$ that corresponds to the $k^{\text{th}}$ class. The classifier $f$ partitions the space $\mathbb{R}^d$ into \textit{classification regions} $\mathcal{R}_1, \dots, \mathcal{R}_{L}$ of constant label. That is, for any $\x \in \mathcal{R}_i$, $\hat{k}(\x) = i$. For a neighboring class $j$, the pairwise decision boundary of the classifier (between these two classes $i$ and $j$) is defined as the set $\mathscr{B}~=~\{~\z:~F(\z) = 0\}$, where $F(\z) = f_{i} (\z) - f_j (\z)$ (we omit dependence on $i$,$j$ for simplicity). The decision boundary defines a hypersurface (of dimension $d-1$) in the $\mathbb{R}^d$. Note that for any point on the decision boundary $\z \in \mathscr{B}$, the gradient $\nabla F(\z)$ is orthogonal to the tangent space $\mathcal{T}_{\z} (\mathscr{B})$ of $\mathscr{B}$ at $\z$.

In this paper, we are interested in studying the decision boundary of a deep neural network in the vicinity of natural images. To do so, for a given point $\x$, we define the mapping $\r(\x)$, given by
$\r(\x) = \argmin_{\r \in \mathbb{R}^d} \| \r \|_2 \text{ subject to } \hat{k} (\x+\r) \neq \hat{k} (\x)$,
which corresponds to the smallest perturbation required to misclassify image $\x$. Note that $\r(\x)$ corresponds geometrically to the vector of minimal norm required to reach the decision boundary of the classifier, and is often dubbed an \textit{adversarial perturbation} \cite{szegedy2013intriguing}. It should further be noted that, due to simple optimality conditions, $\r(\x)$ is orthogonal to the decision boundary at $\x+\r(\x)$. 

In the remainder of this paper, our goal is to analyze the geometric properties of classification regions and decision boundaries of deep networks. In particular, we study the connectedness of classification regions in Sec. \ref{sec:topology}, and the curvature of decision boundaries in Sec. \ref{sec:curvature}, and draw a connection with the robustness of classifiers. We then use the developed geometric insights, and propose a method in Sec. \ref{sec:asymmetry} to detect artificially perturbed data points, and improve the robustness of classifiers.

\section{Topology of classification regions}

\label{sec:topology}

\begin{figure}
  \begin{subfigure}[t]{0.49\textwidth}
    \includegraphics[width=\textwidth,page=2]{path_illustration.pdf}
    \caption{}
    \label{fig:classification_region}
  \end{subfigure}
  \begin{subfigure}[t]{0.49\textwidth}
    \includegraphics[width=\textwidth,page=1]{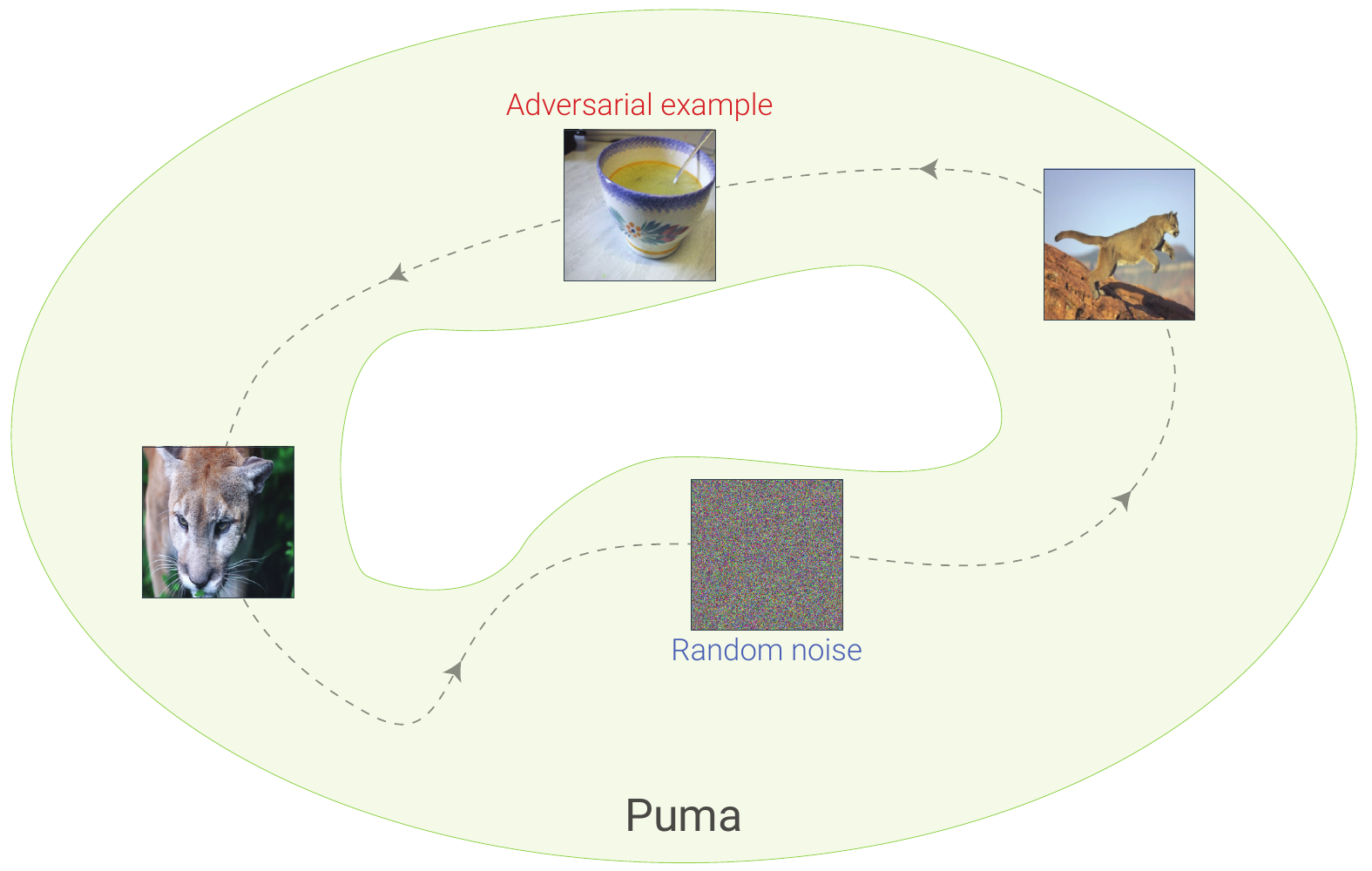}
    \caption{}
    \label{fig:scenarios}
  \end{subfigure}
  \caption{\textbf{(a)} Disconnected versus connected yet complex classification regions. \textbf{(b)} All four images are classified as puma. There exists a path between two images classified with the same label.}
  \label{}
\end{figure}

Do deep networks create shattered and disconnected classification regions, or on the contrary,  one large connected region per label (see Fig. \ref{fig:classification_region})? While deep neural networks have an exponential  number of linear regions (with respect to the number of layers) in the input space \cite{montufar2014number}, it remains unclear whether deep nets create one connected region per class, or shatters a classification region around a large number of small connected sets. We formally cast the problem of connectivity of classification regions as follows: given any two data points $\x_1, \x_2 \in \mathcal{R}_i$, does a continuous curve $\gamma: [0, 1] \rightarrow \mathcal{R}_i$ exist, such that $\gamma(0) = \x_1, \gamma(1) = \x_2$? The problem is complex to address theoretically; we therefore propose a heuristic method to study this question.

To assess the connectivity of regions, we propose a path finding algorithm between two points belonging to the same classification region. That is, given two points $\x_1, \x_2 \in \mathbb{R}^d$, our proposed approach attempts to construct a piecewise linear path $\mathcal{P}$ that remains in the classification region. The path $\mathcal{P}$ is represented as a finite set of anchor points $(\p_0=\x_1, \p_1, \dots, \p_n, \p_{n+1}=\x_2)$, where a convex path is taken between two consecutive points.
To find the path (i.e., the anchor points), the algorithm first attempts to take a convex path between $\x_1$ and $\x_2$; when the path is not entirely included in the classification region, the path is modified by projecting the midpoint $\p = (\x_1+\x_2)/2$ onto the target classification region. The same procedure is applied recursively on the two segments of the path $(\x_1, \p)$ and $(\x_2, \p)$ till the whole path is entirely in the region. The algorithm is summarized in Algorithm \ref{alg:path}.
\begin{algorithm}[h!]
\caption{\label{alg:path}
Finding a path between two data points.}
\begin{algorithmic}[1]
\Function{FindPath}{$\x_1$,$\x_2$}
\State // input: Datapoints $\x_1, \x_2 \in \mathbb{R}^d$.
\State // output: Path $\mathcal{P}$ represented by a set of anchor points. (A convex path is taken between two any anchor point).
  \State $\x_\text{m}\leftarrow(\x_1+\x_2)/2$
  \If{$\hat{k}(\x_\text{m})\ne\hat{k}(\x_1)$}
    \State $\r\leftarrow\argmin_{\r} \|\r\|_2$  $\text{s.t. }\hat{k}(\x_\text{m}+\r)=\hat{k}(\x_1)$
    \State $\x_\text{m}\gets \x_\text{m}+\r$
  \EndIf
  \State $\mathcal{P}\gets(\x_1,\x_\text{m},\x_2)$
  \State // Check the validity of the path by sampling in the convex combinations of consecutive anchor points, and check whether the sampled points belong to region $\hat{k} (\x_1)$.
  \If{$\mathcal{P}$ is a valid path}
    \State \Return $\mathcal{P}$
  \EndIf
  \State $\mathcal{P}_1\gets$\Call{FindPath}{$\x_1$,$\x_\text{m}$}
  \State $\mathcal{P}_2\gets$\Call{FindPath}{$\x_\text{m}$,$\x_2$}
  \State $\mathcal{P}\gets \text{\textbf{concat}}(\mathcal{P}_1,\mathcal{P}_2)$
  \State \Return $\mathcal{P}$
\EndFunction
\end{algorithmic}
\end{algorithm}

We now use the proposed approach to assess the connectivity of the CaffeNet architecture \cite{jia2014} on the ImageNet classification task. To do so, we examine  the existence of paths between
\begin{enumerate}
\item Two randomly sampled points from the validation set with the same estimated label,
\item A randomly sampled point from the validation set, and an adversarially perturbed image \cite{szegedy2013intriguing}. That is, we consider $\x_1$ to be an image from the validation set, and $\x_2 = \tilde{\x}_2 + \r$, where $\tilde{\x}_2$ corresponds to an image classified differently than $\x_1$. $\x_2$ is however classified similarly as $\x_1$, due to the targeted perturbation $\r$.
\item A randomly sampled point from the validation set, and a perturbed random point. This is similar to scenario 2, but $\tilde{\x}_2$ is set to be a random image (i.e., an image sampled uniformly at random from the sphere $\rho \mathbb{S}^{d-1}$, where $\rho$ denotes the typical norm of images).
\end{enumerate}
Note that in scenario $2$ and $3$, $\x_2$ does \textit{not visually} correspond to an image of the same class of $\x_1$ (but is classified by the network as so). These scenarios are illustrated in Fig. \ref{fig:scenarios}. For each scenario, 1,000 pairs of points are considered, and the approach described above is used to find the path. Our result can be stated as follows:

\begin{center}
\textit{In all three scenarios, a continuous path always exists between points sampled from the same classification region.}
\end{center}

\begin{figure}[ht!]
  \centering
  \includegraphics[width=\textwidth]{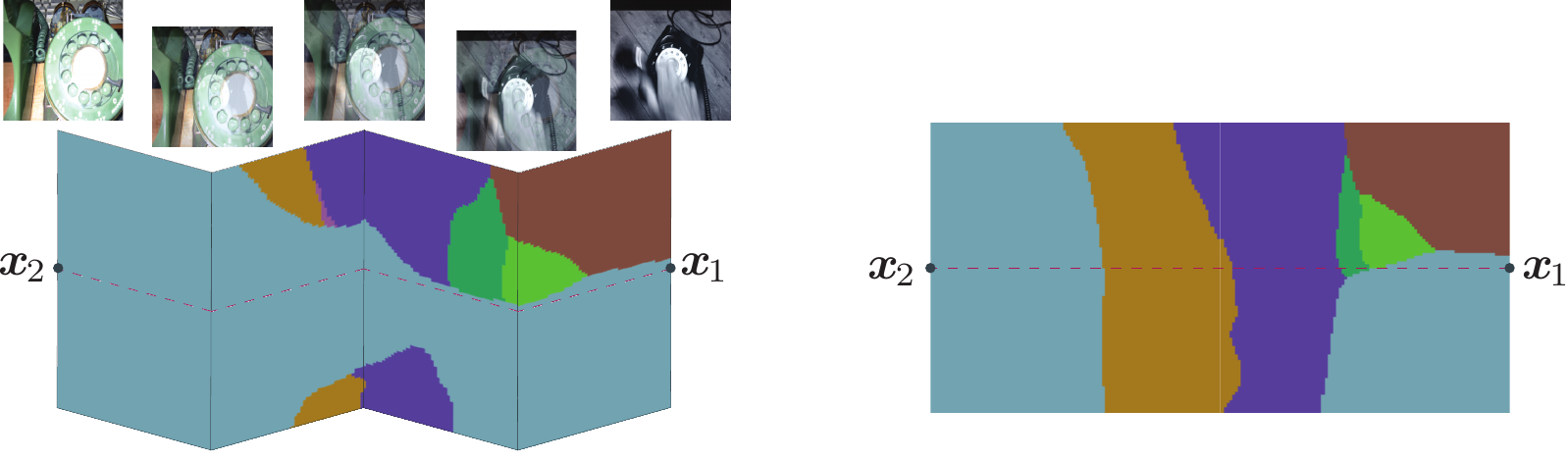}
  \caption{Classification regions (shown with different colors), and illustration of different paths between images $\x_1$, $\x_2$. \textbf{Left:} The convex path between two datapoints might not be entirely included in the classification region (note that the linear path traverses $4$ other regions). The image is the cross-section spanned by $\r(\x_1)$ (adversarial perturbation of $\x_1$) and $\x_1-\x_2$. \textbf{Right:} Illustration of a nonconvex path that remains in the classification region. The image is obtained by stitching cross-sections spanned by $\r(\x_1)$ and $\p_i-\p_{i+1}$ (two consecutive anchor points in the path $\mathcal{P}$).}
  \label{fig:panorama}
\end{figure}
\begin{figure}[ht!]
  \centering
\includegraphics[width=0.4\textwidth]{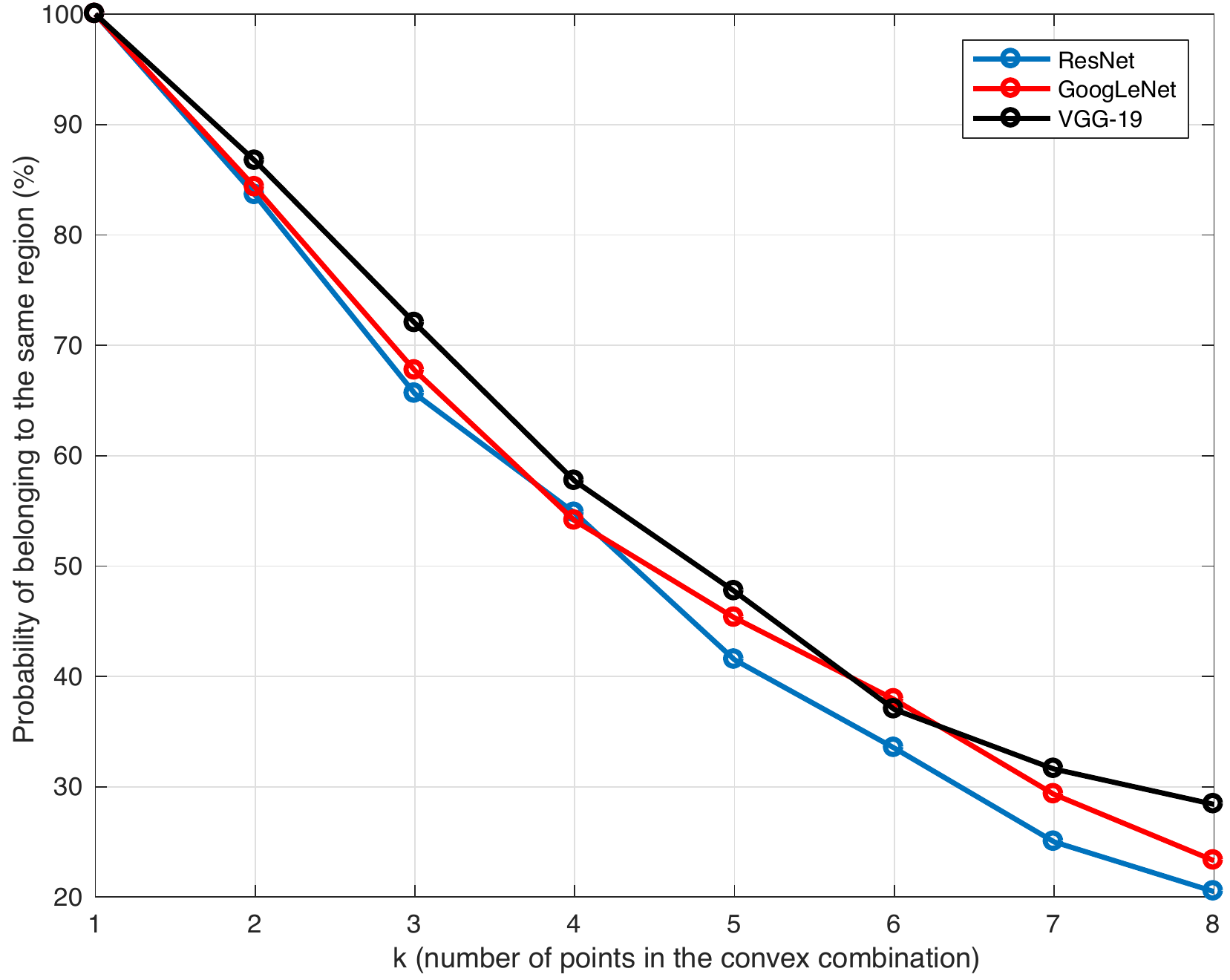}
\caption{\label{fig:convex_hull_exp} Empirical probability ($y$ axis) that a convex combination of $k$ samples ($x$ axis) from the same classification region stays in the classification region, for networks trained on ImageNet. Samples are randomly chosen from the validation set.}
\end{figure}
This result suggests that the classification regions created by deep neural networks are \textit{connected} in $\mathbb{R}^d$: deep nets create single large regions containing all points of the same label.  
More than that, the path that was found using the proposed path finding approach approximately corresponds to a straight path. An illustration of the path between two images from the validation set (i.e. scenario $1$) is provided in Fig. \ref{fig:panorama}.

Interestingly, when the endpoints are two randomly sampled images from the validation set, the straight path between the two endpoints overwhelmingly belong to the classification region. However, classification regions are \textit{not} convex bodies in $\mathbb{R}^d$. Fig.~\ref{fig:convex_hull_exp} illustrates the estimated probability that random convex combinations of $k$ images $\x_1, \dots, \x_k \in \mathcal{R}_i$ belong to $\mathcal{R}_i$. Observe that while convex combinations of two samples in a region are very likely (with probability $\approx 80\%$) to belong to the same region, convex combinations of $\geq 5$ samples do \textit{not} usually belong to the same region.

Our experimental results therefore suggest that deep neural networks create large \textit{connected} classification regions, where any two points in the region are connected by a path.

In the next section, we explore the \textit{complexity} of the boundaries of these classification regions learned by deep networks, through their curvature property.\\\\
\section{Curvature of the decision boundaries}
\label{sec:curvature}
\begin{figure}
  \center
  \begin{subfigure}{0.49\textwidth}
    \center
  \includegraphics[width=0.8\textwidth]{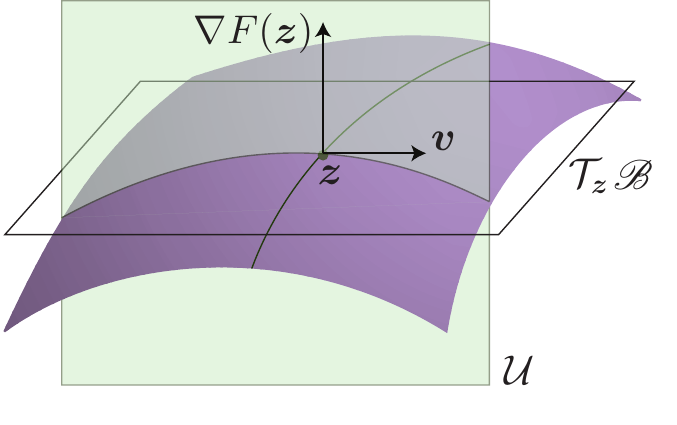}
  \caption{}
  \label{fig:geometric_fig}
  \end{subfigure}
  ~
  \begin{subfigure}{0.49\textwidth}
    \center
  \includegraphics[width=0.8\textwidth]{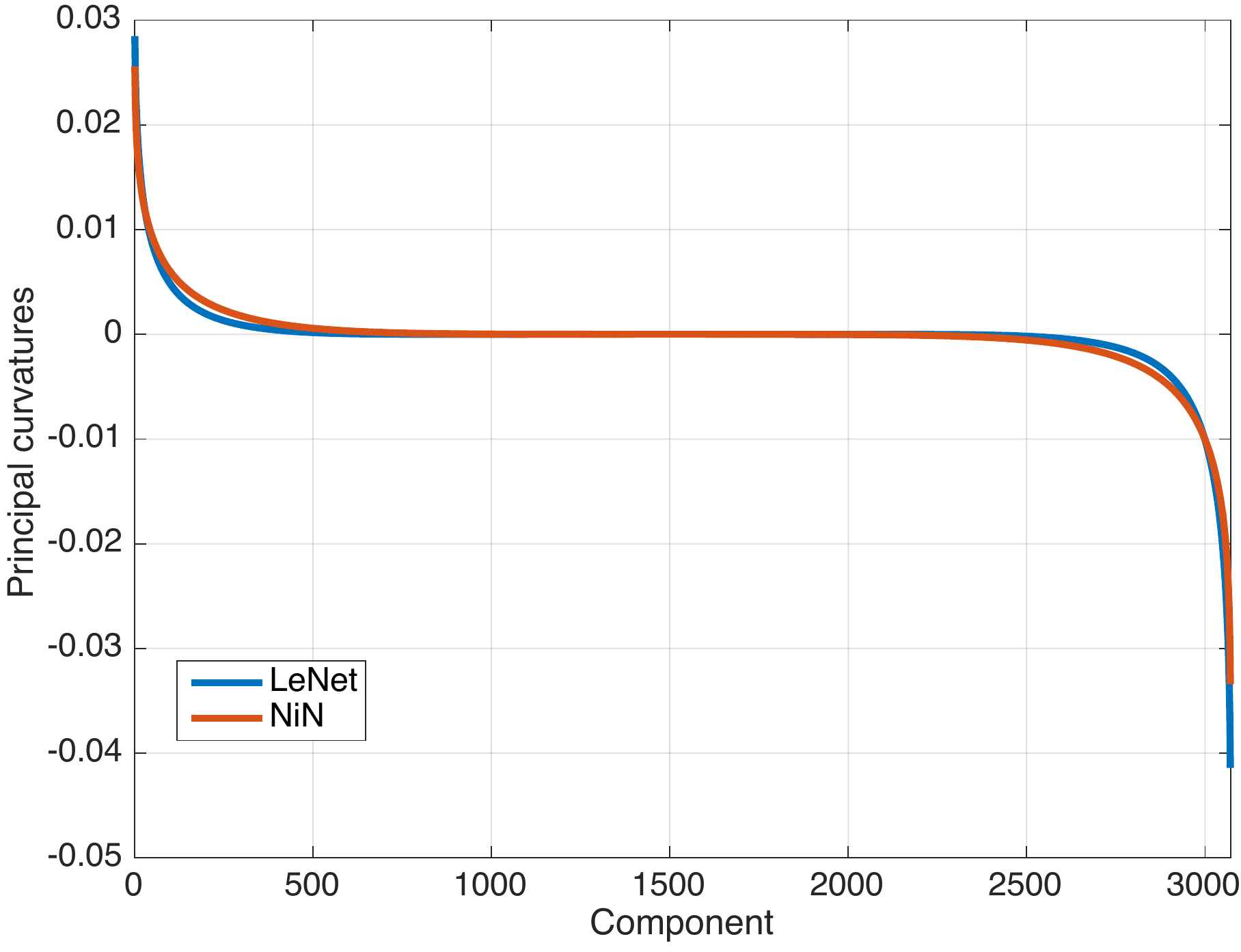}
  \caption{}
  \label{fig:eigs_profiles}
  \end{subfigure}
  \caption{\label{fig:geom_eigs} \textbf{(a)} Normal section $\mathcal{U}$ of the decision boundary, along the plane spanned by the normal vector $\nabla F(\z)$ and $\v$. \textbf{(b)} Principal curvatures for NiN and LeNet networks, computed at a point $\z$ on the decision boundary in the vicinity of a natural image. }
\end{figure}

We start with basic definitions of curvature. 
The \textit{normal} curvature $\kappa(\z, \v)$ along a tangent direction $\v \in \mathcal{T}_{\z} (\mathscr{B})$ is defined as the curvature of the planar curve resulting from the cross-section of $\mathscr{B}$ along the two-dimensional normal plane spanning $(\nabla F(\z), \v)$ (see Fig. \ref{fig:geometric_fig} for details). The curvature along a tangent vector $\v$ can be expressed in terms of the Hessian matrix $H_F$ of $F$ \cite{lee2009manifolds}:
 \begin{equation}
 \kappa(\z, \v) = \frac{\v^T H_F \v}{\| \v \|_2^2 \| \nabla F(\z) \|_2}.
 \label{eq:curvature}
 \end{equation}
 Principal directions correspond to the orthogonal directions in the tangent space maximizing the curvature $\kappa(\z, \v)$. Specifically, the $l$-th principal direction $\v_l$ (and the corresponding principal curvature $\kappa_l$) is obtained by maximizing $\kappa(\z,\v)$ with the constraint $\v_l \perp \v_{1} \dots \v_{l-1}$. Alternatively, the principal curvatures correspond to the nonzero eigenvalues of the matrix $\frac{1}{\| \nabla F(\z) \|_2} P H_F P$, where $P$ is the projection operator on the tangent space; i.e., $P = I - \nabla F(\z) \nabla F(\z)^T$. 

We now analyze the curvature of the decision boundary of deep neural networks in the vicinity of natural images. We consider the LeNet and NiN  \cite{lin2013} architectures trained on the CIFAR-10 task, and show the principal curvatures of the decision boundary, in the vicinity of 1,000 randomly chosen images from the validation set. Specifically, for a given image $\x$, the perturbed sample $\z = \x+\r(\x)$  corresponds to the closest point to $\x$ on the decision boundary. We then compute the principal curvatures at point $\z$ with Eq. \ref{eq:curvature}. The average profile of the principal curvatures (over $1,000$ data points) is illustrated in Fig. \ref{fig:eigs_profiles}.
Observe that, for both networks, the large majority of principal curvatures are approximately zero: along these principal directions, the decision boundary is almost flat. Along the remaining principal directions, the decision boundary has (non-negligible) positive or negative curvature.
Interestingly, the principal curvature profile is asymmetric towards \textit{negatively curved} directions. This property is \textit{not} specific to the considered datapoints, the considered networks, or even the problem at hand (e.g., CIFAR-10). In fact, this bias towards negatively curved directions is repeatable across a wide range of networks and datasets. In the next section, we leverage this characteristic asymmetry of the decision boundaries of deep neural networks (in the vicinity of natural images) in order to detect adversarial examples from clean examples.

\begin{figure}[t]
\centering
\begin{subfigure}[t]{0.40\textwidth}
  \includegraphics[width=\textwidth]{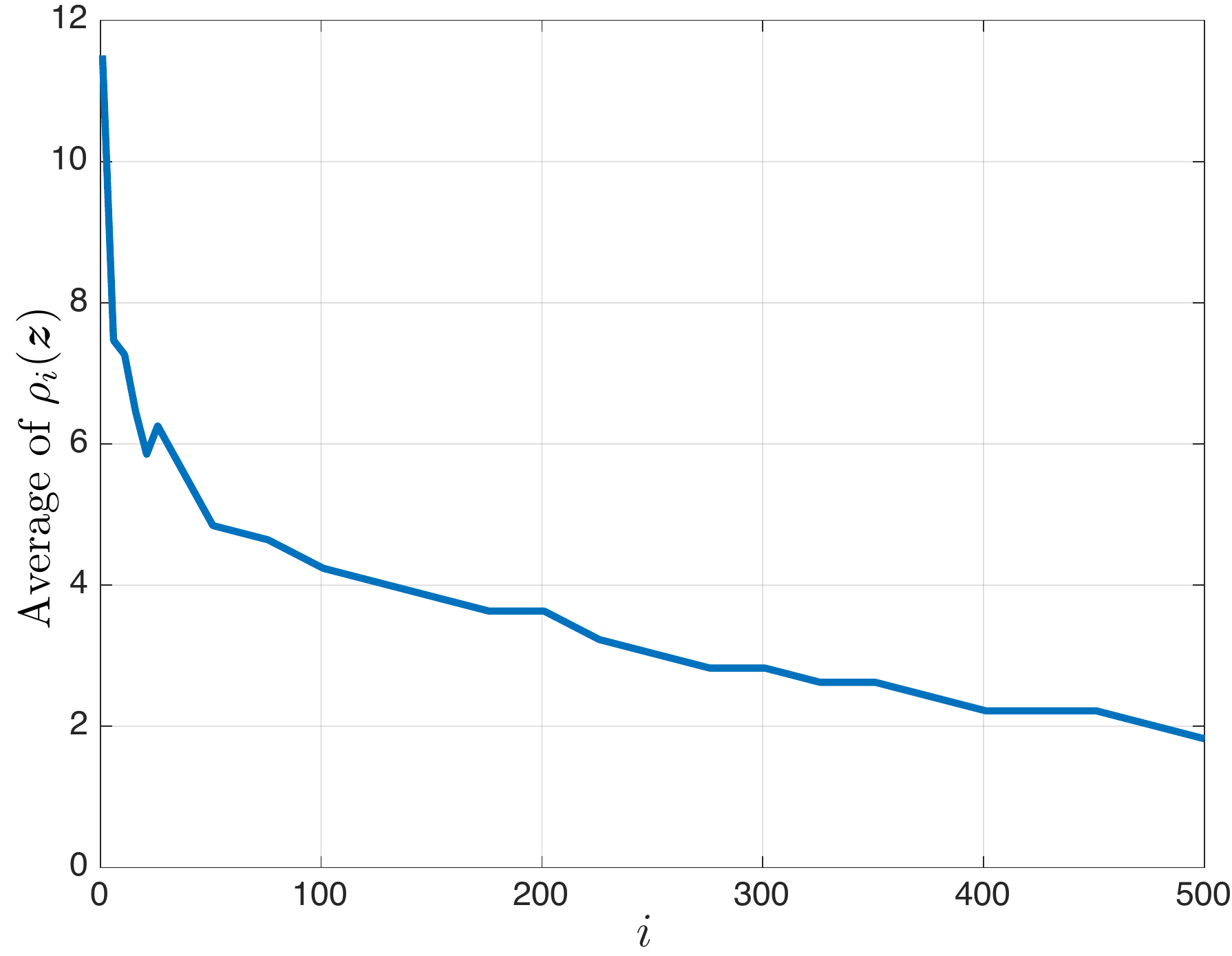}
  \caption{\label{fig:rho_S}}
\end{subfigure}
\begin{subfigure}[t]{0.40\textwidth}
  \centering
  \includegraphics[width=0.7\textwidth]{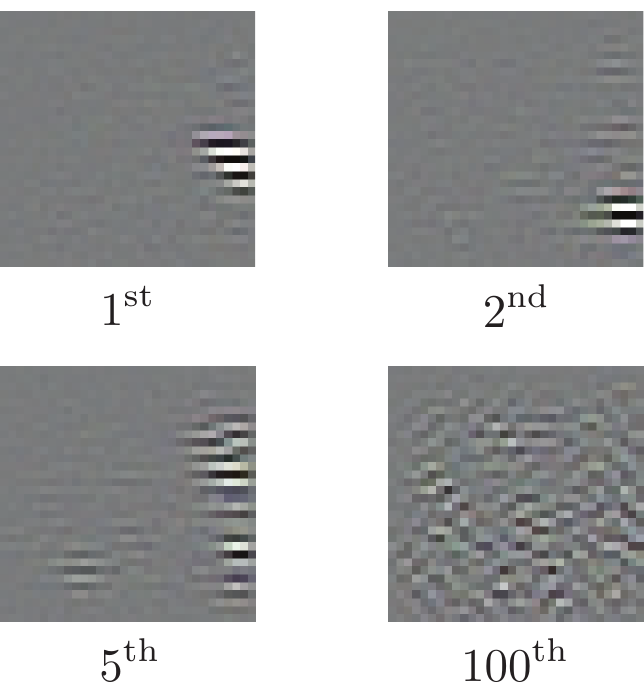}
  \caption{\label{fig:basis}}
\end{subfigure}
\caption{\label{fig:fig_5} \textbf{(a)} Average of $\rho_i(\z)$ as a function of $i$ for many different points $\z$ in the vicinity of natural images. \textbf{(b)} Basis of $\mathcal{S}$.}
\end{figure}


While the above local analysis shows the existence of few directions along which the decision boundary is curved, we now examine whether these directions are  \textit{shared} across different datapoints, and relate these directions with the robustness of deep nets. 
To estimate the \textit{shared} common curved directions, we compute the largest \textit{principal directions} for a randomly chosen batch of $100$ training samples and merge these directions into a matrix $M$. We then estimate the common curved directions as the $m$ largest singular vectors of $M$ that we denote by $\u_1, \dots, \u_m$. To assess whether the decision boundary is curved in such directions, we then evaluate the curvature of the decision boundary in such directions for points $\z$ in the vicinity of \textit{unseen} samples from the \textit{validation} set. That is, for $\x$ in the validation set, and $\z = \x + \r(\x)$, we compute
$\rho_{i}(\z) = \frac{|\u_i^T P H_F P \u_i|}{\ex{\v \sim \mathbb{S}^{d-1}}{|\v^T P H_F P \v|}}$,
which measures how relatively curved is the decision boundary in direction $\u_i$, compared to random directions sampled from the unit sphere in $\mathbb{R}^d$. 
When $\rho_{i} (\z) \gg 1$, this indicates that $\u_i$ constitutes a direction that significantly curves the decision boundary at $\z$.
Fig. \ref{fig:rho_S} shows the average of $\rho_{i} (\z)$ over 1,000 points $\z$ on the decision boundary in the vicinity of \textit{unseen} natural images, for the LeNet architecture on CIFAR-10. Note that the directions $\u_i$ (with $i$ sufficiently small) lead to universally curved directions across \textit{unseen} points. That is, the decision boundary is highly curved along such  data-independent directions.
Note that, despite using a relatively small number of samples (i.e., $100$ samples) to compute the shared directions , these generalize well to unseen points. We illustrate in Fig. \ref{fig:basis} these directions $\u_i$, along which decision boundary is universally curved in the vicinity of natural images; interestingly, the first principal directions (i.e., directions along which the decision boundary is highly curved) are very localized Gabor-like filters. Through discriminative training, the deep neural network has implicitly learned to curve the decision boundary along such directions, and preserve a flat decision boundary along the orthogonal subspace.

 \begin{figure}[t]
 \begin{floatrow}
 \ffigbox[1.3\FBwidth]{\includegraphics[scale=0.3]{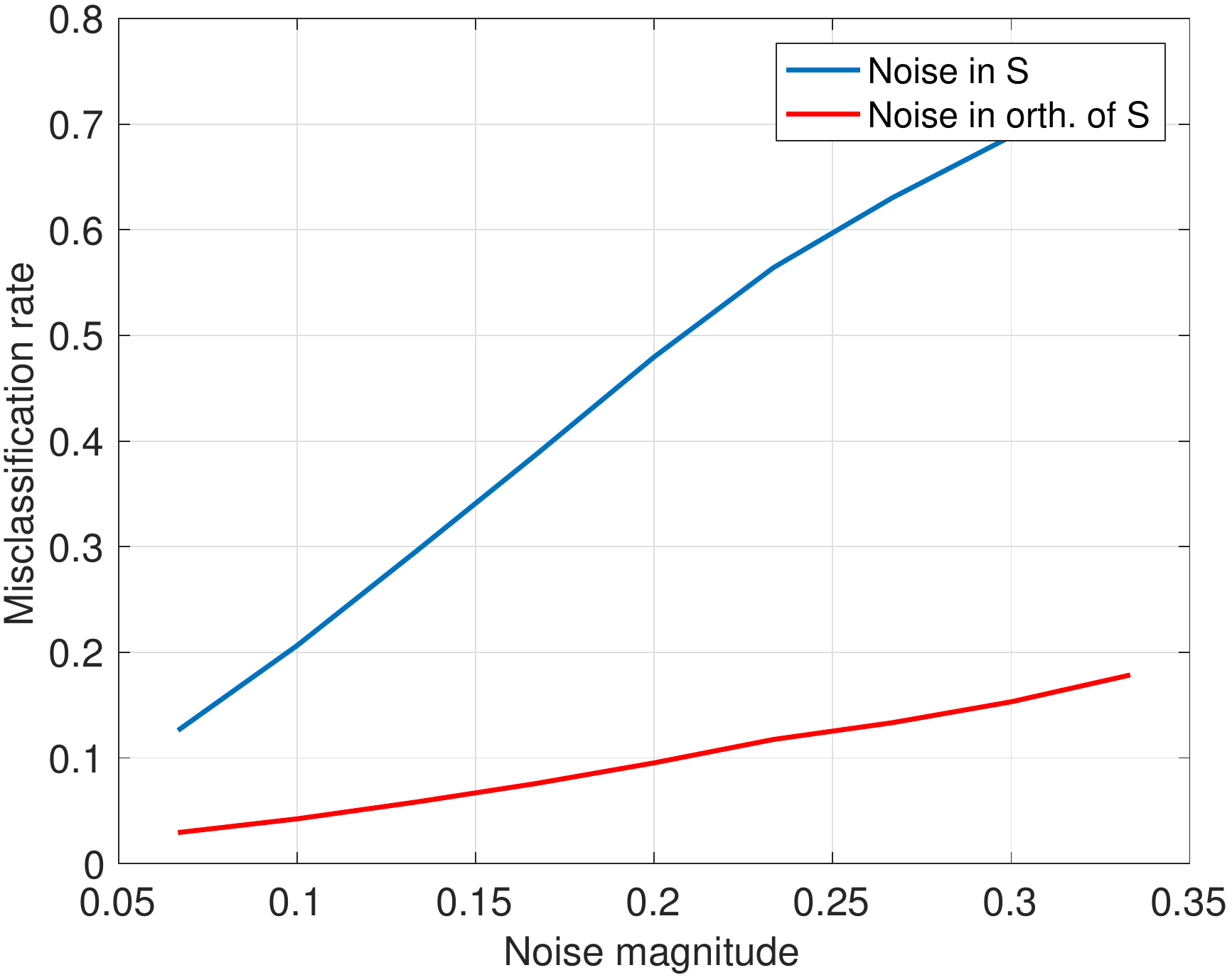}}{
\caption{\label{fig:robustness_curvature} Misclassification rate ($\%$ of images that change labels) on the noisy validation set, with respect to the noise magnitude ($\ell_2$ norm of noise divided by the typical norm of images). }}
 \capbtabbox[1.3\FBwidth]{
\begin{tabular}{|l|c|c|}
\hline
& LeNet & NiN \\ \hline
Random & 0.25 & 0.25 \\ \hline
$\x_2 - \x_1$ & 0.10 & 0.09 \\ \hline
$\nabla \x$ & 0.22 & 0.24 \\ \hline
Adversarial & 0.64 & 0.60 \\ \hline
\end{tabular}}{
\caption{\label{tab:projection_norm} Norm of projected perturbation on $\mathcal{S}$: $\frac{\| P_{\mathcal{S}} \v \|_2}{\| \v \|_2}$. Larger values indicate that perturbation belongs to subspace $\mathcal{S}$.}
 }
 \end{floatrow}
 \end{figure}

Interestingly, the data-independent directions $\u_i$ (where the decision boundary is highly curved) are also tightly connected with the \textit{invariance} of the classifier to  perturbations. To elucidate this relation, we construct a subspace $\mathcal{S} = \text{span} (\u_1, \dots, \u_{200})$ that contains the first $200$ shared curved directions. Then, we show in Fig. \ref{fig:robustness_curvature} the accuracy of the CIFAR-10 LeNet model on a noisy validation set, where the noise either belongs to $\mathcal{S}$, or to $\mathcal{S}^{\perp}$ (i.e., orthogonal of $\mathcal{S}$). It can be observed the deep network is much more robust to noise orthogonal to $\mathcal{S}$, than to noise in $\mathcal{S}$. Hence, $\mathcal{S}$ also represents the subspace of perturbations to which the classifier is highly vulnerable, while the classifier has learned to be invariant to perturbations in $\S^{\perp}$. To support this claim, we report in Table \ref{tab:projection_norm}, the norm of the projection of adversarial perturbations (computed using the method in \cite{moosavi2015deepfool}) on the subspace $\mathcal{S}$, and compare it to that of the projection of random noise onto $\mathcal{S}$. Note that for both networks under study, adversarial perturbations project well onto the subspace $\mathcal{S}$ comparatively to random perturbations, which have a significant component in $\mathcal{S}^{\perp}$. 
In contrast, note that perturbations obtained by taking the difference of two random images belong overwhelmingly to $\mathcal{S}^{\perp}$, which agrees with the observation drawn in Sec. \ref{sec:topology} whereby straight paths are likely to belong to the classification region. Finally, note that the gradient of the image $\nabla \x$ also does not have an important component in $\mathcal{S}$, as the robustness to such directions is fundamental to achieve invariance to small geometric deformations.\footnote{In fact, a first order Taylor approximation of a translated image $\x(\cdot + \tau_1, \cdot + \tau_2) \approx \x + \tau_1 \nabla_x \x + \tau_2 \nabla_y \x$. To achieve robustness to translations, a deep neural network hence needs to be locally invariant to perturbations along the gradient directions.}

The importance of the shared directions $\{ \u_i \}$, where the decision boundary is curved hence goes beyond our curvature analysis, and capture the modes of sensitivity learned by the deep network. 

\section{Exploiting the asymmetry to detect perturbed samples}

\label{sec:asymmetry}
\begin{figure}[t]
 \begin{floatrow}
 \ffigbox{\includegraphics[scale=0.5]{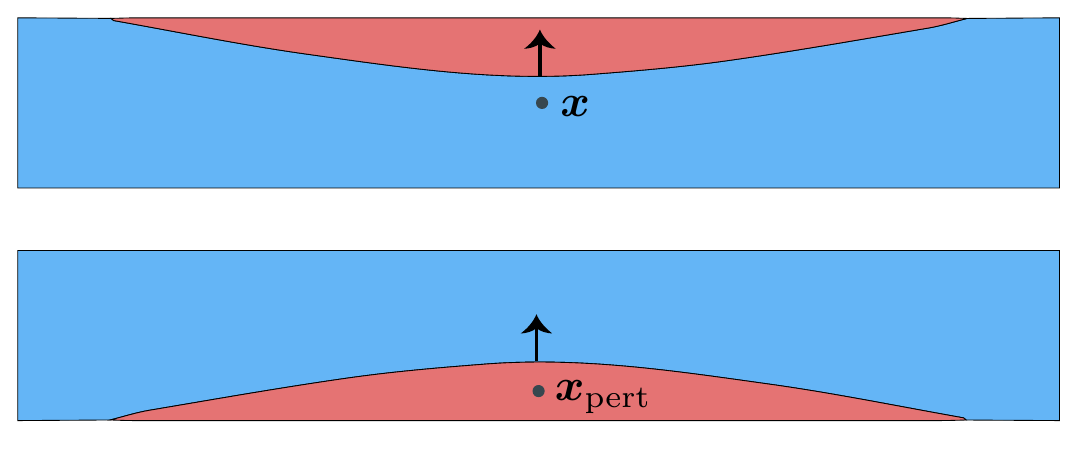}}{
 \caption{\label{fig:schematic_pos_neg}Schematic representation of normal sections in the vicinity of a natural image (top), and perturbed image (bottom). The normal vector to the decision boundary is indicated with an arrow.}}
 \capbtabbox{
 \begin{tabular}{|l|l|l|}
 \hline
  & LeNet & NiN \\ \hline
 $\%$ $\overline{\kappa} > 0$ for original samples & $97\%$ & $94\%$ \\ \hline
 $\%$ $\overline{\kappa} < 0$ for perturbed samples & $96\%$ & $93\%$ \\ \hline
 \end{tabular}}{
 \caption{\label{tab:percentage}Percentage of points on the decision boundary with positive (resp. negative) average curvature, when sampled in the vicinity of natural images (resp. perturbed images).}
 }
 \end{floatrow}
 \end{figure}

In this section, we leverage the asymmetry of the principal curvatures (illustrated in Fig. \ref{fig:eigs_profiles}), and propose a method to distinguish between original images, and images perturbed with adversarial examples, as well as improve the robustness of classifiers.  For an element $\z$ on the decision boundary, denote by $\overline{\kappa}(\z) = \frac{1}{d-1} \sum_{i=1}^{d-1} \kappa_i (\z)$ the average of the principal curvatures. For points $\z$ sampled in the vicinity of natural images, the profile of the principal curvature is asymmetric (see Fig. \ref{fig:eigs_profiles}), leading to a negative average curvature; i.e., $\overline{\kappa}(\z) < 0$. In contrast, if
$\x$ is now perturbed with an adversarial example (that is, we observe $\x_{\text{pert}} = \x + \r(\x)$ instead of $\x$), the average curvature at the vicinity of $\x_{\text{pert}}$ is instead \textit{positive}, as schematically illustrated in Fig. \ref{fig:schematic_pos_neg}. Table \ref{tab:percentage} supports this observation empirically with adversarial examples computed with the method in \cite{moosavi2015deepfool}. Note that for both networks, the asymmetry of the principal curvatures allows to distinguish very accurately original samples from perturbed samples using the \textit{sign} of the curvature. Based on this simple idea, we now derive an algorithm for detecting adversarial perturbations.

Since the computation of all the principal curvatures is intractable for large-scale datasets, we now derive a tractable estimate of the average curvature.
Observe that the average curvature $\overline{\kappa}$ can be equivalently written as $\ex{v \sim \mathbb{S}^{d-1}}{\v^T G(\z) \v}$,
where $G(\z) = \| \nabla F(\z) \|_2^{-1} (I-\nabla F(\z) \nabla F(\z)^T) H_F (\z) (I-\nabla F(\z) \nabla F(\z)^T)$. In fact, we have
\[
\ex{\v \sim \mathbb{S}^{d-1}}{\v^T G(\z) \v} = \ex{\v \sim \mathbb{S}^{d-1}}{\v^T \left( \sum_{i=1}^{d-1} \kappa_i \v_i \v_i^T \right) \v} = \frac{1}{d-1} \sum_{i=1}^{d-1} \kappa_i,
\]
where $\v_i$ denote the principal directions. It therefore follows that the average curvature $\overline{\kappa}$ can be efficiently estimated using a sample estimate of $\ex{v \sim \mathbb{S}^{d-1}}{\v^T G(\z) \v}$ (and without requiring the full eigen-decomposition of $G$). To further make the approach of detecting perturbed samples more practical, we approximate $G(\z)$ (for $\z$ on the decision boundary) with $G(\x)$, assuming that $\x$ is sufficiently close to the decision boundary.\footnote{The matrix $G$ is never computed in practice, since only matrix vector multiplications of $G$ are needed.} This approximation avoids the computation of the closest point on the decision boundary $\z$, for each $\x$.

\begin{figure}
  \begin{subfigure}{0.5\textwidth}
    \center
    \includegraphics[width=0.8\textwidth]{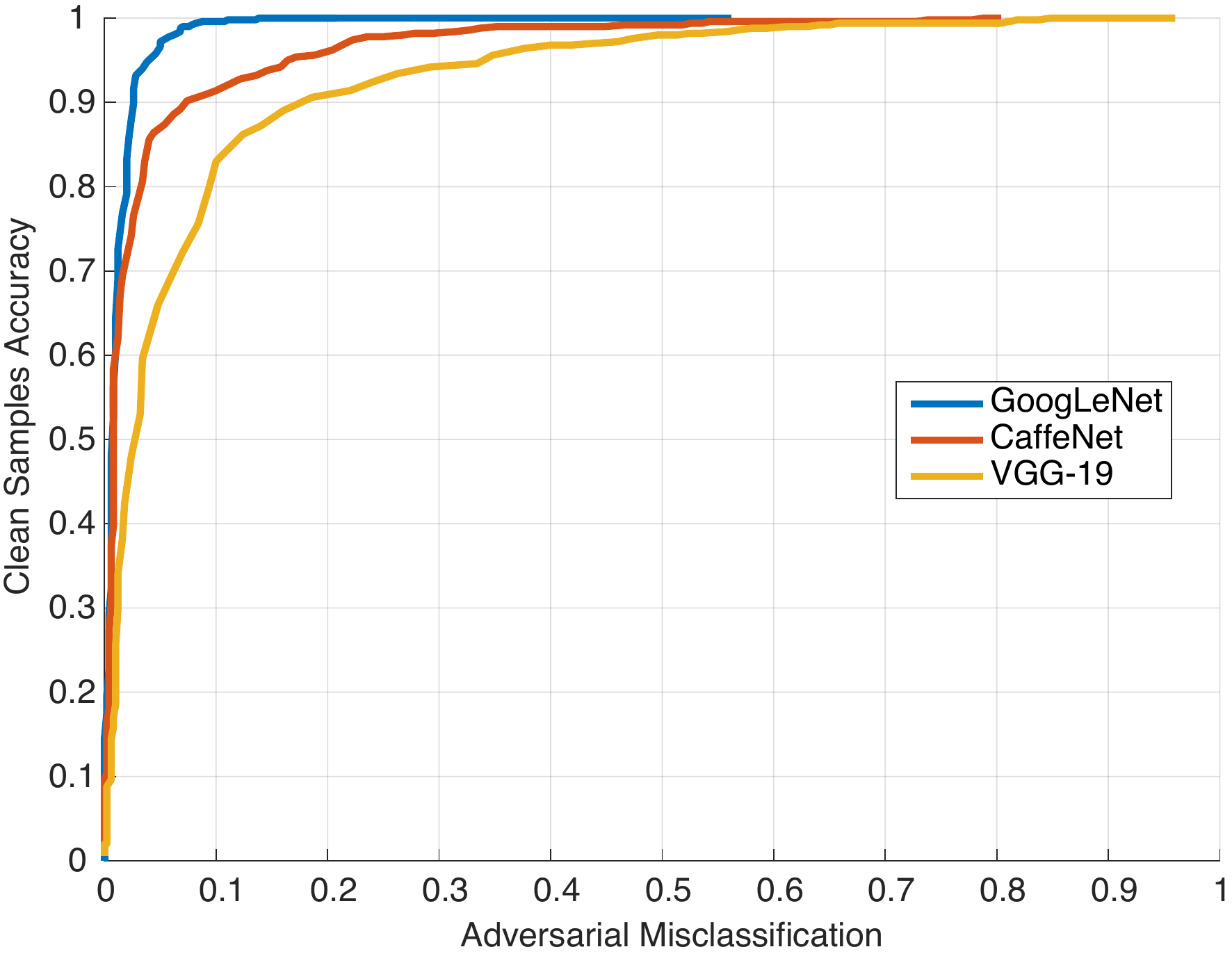}
  \end{subfigure}%
  \begin{subfigure}{0.5\textwidth}
    \center
    \includegraphics[width=0.8\textwidth]{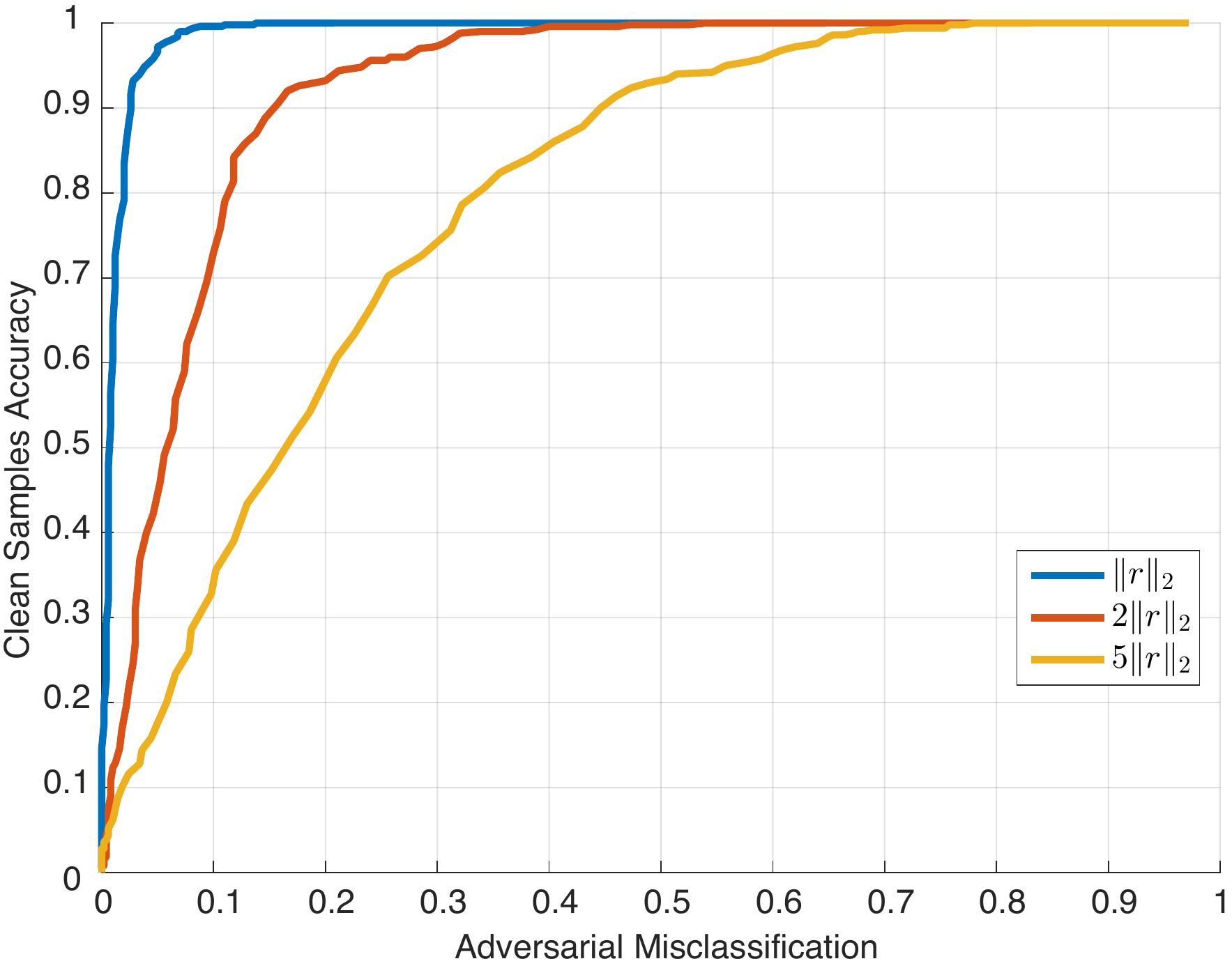}
  \end{subfigure}
  \caption{True positives (i.e., detection accuracy on \textit{clean samples}) vs. False positives (i.e., detection error on \textit{perturbed samples}). \textbf{Left:} Results reported for GoogLeNet, CaffeNet and VGG-19 architectures, with perturbations computed using the approach in \cite{moosavi2015deepfool}. \textbf{Right:} Results reported for GoogLeNet, where perturbations are scaled by a constant factor $\alpha = 1, 2, 5$.}
  \label{fig:detecting_perturbed}
\end{figure}

\begin{algorithm}[ht]
\caption{\label{alg:detect}
Detecting and denoising perturbed samples.}
\begin{algorithmic}[1]
\State \textbf{input: } classifier $f$, sample $\x$, threshold $t$.
\State \textbf{output: } boolean \textit{perturbed}, recovered label \textit{label}.
\State Set $F_i \gets f_i - f_{\hat{k}}$ for $i \in [L]$.
\State Draw iid samples $\v_1, \dots, \v_T$ from the uniform distribution on $\mathbb{S}^{d-1}$.
\State Compute
$\displaystyle \rho \gets \frac{1}{L T} \sum_{\substack{i=1 \\ i \neq \hat{k}(\x)}}^{L} \sum_{j=1}^{T} \v_j^T G_{F_i} \v_j,$
where $G_{F_i}$ denotes the Hessian of $F_i$ projected on the tangent space; i.e., $G_{F_i} (\x) = \| \nabla F(\x) \|_2^{-1} (I - \nabla F(\x) \nabla F(\x)^T) H_{F_i} (\x) (I - \nabla F(\x) \nabla F(\x)^T).$
\If{$\rho < t$} $\textit{perturbed} \gets \textit{false}$.
\Else \hspace{1mm} $\textit{perturbed} \gets \textit{false}$ and $\textit{label} \gets \underset{{\substack{i \in \{1, \dots, L\} \\ i \neq \hat{k} (\x)}}}{\argmax} \sum_{j=1}^T \v_j^T G_{F_i} \v_j$.
%
\EndIf
\end{algorithmic}
\end{algorithm}

We provide the details in Algorithm \ref{alg:detect}. Note that, in order to extend this approach to multiclass classification, an empirical average is taken over the decision boundaries with respect to all other classes. Moreover, while we have used a threshold of $0$ to detect adversarial examples from original data in the above explanation, a threshold parameter $t$ is used in practice (which controls the true positive vs. false positive tradeoff). Finally, it should be noted that in addition to detecting whether an image is perturbed, the algorithm also provides an estimate of the original label when a perturbed sample is detected (the class leading to the highest positive curvature is returned).

We now test the proposed approach on different networks trained on ImageNet, with adversarial examples computed using the approach in \cite{moosavi2015deepfool}. The latter approach is used as it provides small and difficult to detect adversarial examples, as mentioned in \cite{metzen2017detecting, lu2017safetynet}. Fig. \ref{fig:detecting_perturbed} (left) shows the accuracy of the detection of Algorithm \ref{alg:detect} on \textit{original} images with respect to the detection error on \textit{perturbed} images, for varying values of the threshold $t$. For the three networks under test, the approach achieves very accurate detection of adversarial examples (e.g., more than $95\%$ accuracy on GoogLeNet with an optimal threshold). Note first that the success of this strategy confirms the asymmetry of the curvature of the decision boundary on the more complex setting of large-scale networks trained on ImageNet. Moreover, this simple curvature-based detection strategy outperforms the detection approach recently proposed in \cite{lu2017safetynet}. In addition, unlike other approaches of detecting perturbed samples (or improving the robustness), our approach only uses the characteristic geometry of the decision boundary of deep neural networks (i.e., the curvature \textit{asymmetry}), and does not involve any training/fine-tuning with perturbed samples, as commonly done. 

The proposed approach not only distinguishes original from perturbed samples, but it also provides an estimate of the correct label, in the case a perturbed sample is detected. Algorithm \ref{alg:detect} correctly recovers the labels of perturbed samples with an accuracy of $92\%$, $88\%$ and $74\%$ respectively for GoogLeNet, CaffeNet and VGG-19, with $t = 0$. This shows that the proposed approach can be effectively used to denoise the perturbed samples, in addition to their detection.

Finally, we report in Fig. \ref{fig:detecting_perturbed} (right) reports a similar graph to that of Fig. \ref{fig:detecting_perturbed} (left) for the GoogLeNet architecture, but where the perturbations are now multiplied by a factor $\alpha \geq 1$. Note that, as $\alpha$ increases, the accuracy of detection using of our method decreases, as it heavily relies on \textit{local} geometric properties of the classifier (i.e., the curvature). Interestingly enough, \cite{lu2017safetynet, metzen2017detecting} report that the regime where perturbations are very small (like those produced by \cite{moosavi2015deepfool}) are the hardest to detect; we therefore foresee that this geometric approach will be used along with other detection approaches, as it provides very accurate detection in a distinct regime where traditional detectors do not work well (i.e., when the perturbations are very small).

\vspace{-3mm}
\section{Conclusion}

We analyzed in this paper the geometry induced by deep neural network classifiers in the input space. Specifically, we provided empirical evidence showing that classification regions are connected. Next, to analyze the complexity of the functions learned by deep networks, we provided a comprehensive empirical analysis of the curvature of the decision boundaries. We showed in particular that, in the vicinity of natural images, the decision boundaries learned by deep networks are flat along most (but not all) directions, and that some curved directions are \textit{shared} across datapoints. We finally leveraged a fundamental observation on the \textit{asymmetry} in the curvature of deep nets, and proposed an algorithm for detecting adversarially perturbed samples from original samples. This geometric approach was shown to be very effective, when the perturbations are sufficiently small, and that recovering the label was further possible using this algorithm. This shows that the study of the geometry of state-of-the-art deep networks is not only key from an analysis (and understanding) perspective, but it can also lead to classifiers with better properties.
\subsubsection*{Acknowledgments}
S.M and P.F gratefully acknowledge the support of NVIDIA Corporation with the donation of the Titan X Pascal GPU used for this research. This work has been partly supported by the Hasler Foundation, Switzerland, in the framework of the ROBERT project. A.F was supported by the Swiss National
Science Foundation under grant P2ELP2-168511. S.S. was supported by ONR N00014-17-1-2072 and ARO W911NF-15-1-0564.

\small{
\bibliographystyle{ieeetr}
\bibliography{bibliography}
}

\end{document}